\documentclass[10pt,twocolumn,letterpaper]{article}

\usepackage{iccv}
\usepackage{times}
\usepackage{epsfig}
\usepackage{graphicx}
\usepackage{amsmath}
\usepackage{amssymb}

\usepackage{amssymb}
\usepackage{pifont}
\usepackage{booktabs} 
\usepackage{xcolor}
\usepackage{multirow}
\usepackage{subfig}
\usepackage{nicefrac}
\usepackage[normalem]{ulem}
\useunder{\uline}{\ul}{}
\DeclareMathOperator*{\argmin}{arg\,min}

\usepackage[pagebackref=true,breaklinks=true,letterpaper=true,colorlinks,bookmarks=false]{hyperref}

\iccvfinalcopy 

\newcommand{\student}[0]{\textit{student}}
\newcommand{\teacher}[0]{\textit{teacher}}

\newcommand{\cmark}{\ding{51}}%
\newcommand{\xmark}{\ding{55}}%

\def\iccvPaperID{17} 

\ificcvfinal\fi

\begin{document}

\title{Temporal DINO: A Self-supervised Video Strategy to Enhance Action Prediction}

\author{Izzeddin Teeti$^{1, *}$, Rongali Sai Bhargav$^{2}$, Vivek Singh$^{1}$, Andrew Bradley$^{1}$, \\
Biplab Banerjee$^{2}$, Fabio Cuzzolin$^{1}$
\\
\small{
$^1$VAIL, Oxford Brookes University, (iteeti, vsingh, abradley, fabio.cuzzolin)@brookes.ac.uk,} \\
\small{$^2$Indian Institute of Technology, Bombay, (sai.bhargav, Biplab)@iitb.ac.in}
}

\maketitle
\ificcvfinal\thispagestyle{empty}\fi

\vspace{-0.5cm}
\begin{figure*}[t]
    \centering
    \includegraphics[width=\linewidth]{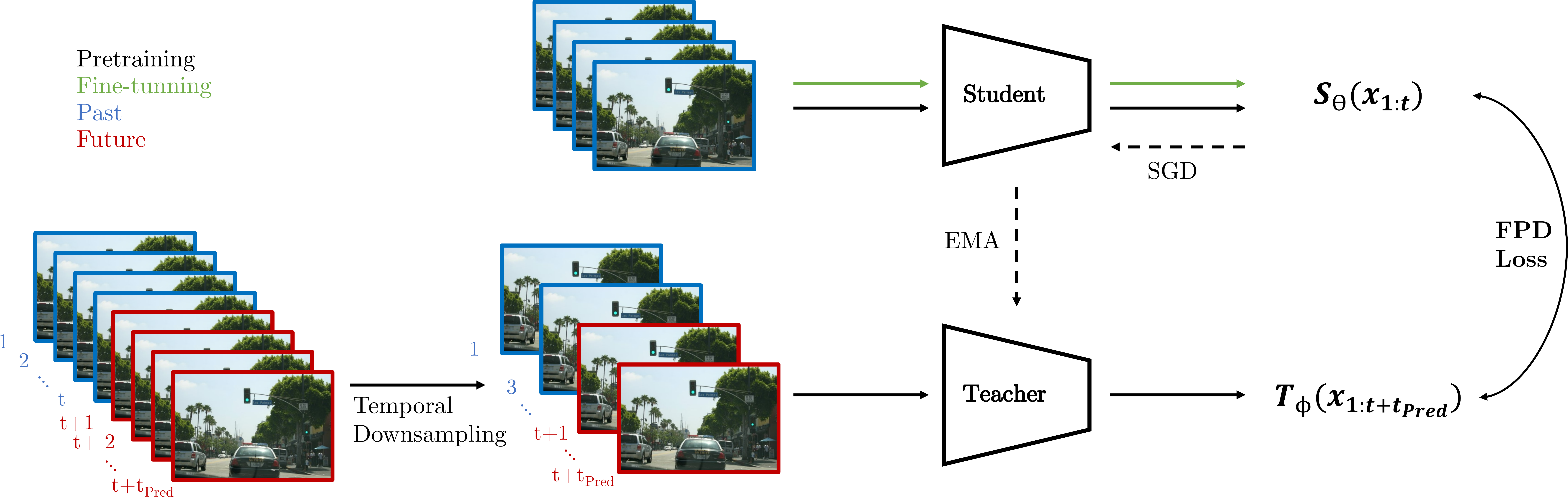}
    \caption{\small Overview of the proposed Temporal DINO. The \student~ model processes the \textcolor{blue}{past} frames ($x_{1:t}$), while the \teacher~ processes both the \textcolor{blue}{past} and \textcolor{red}{future} frames ($x_{1:t+t_{Pred}}$)
    . A Future-past Distillation loss is applied to their representations ($S_{\theta}$ and $T_{\phi}$) to guide the \student~ to capture the future temporal context from the \teacher. 
    }
    \label{fig:overview}
\end{figure*}

\begin{abstract}
The emerging field of action prediction - the task of forecasting action in a video sequence - plays a vital role in various computer vision applications such as autonomous driving, activity analysis and human-computer interaction. Despite significant advancements, accurately predicting future actions remains a challenging problem due to high dimensionality, complex dynamics and uncertainties inherent in video data. Traditional supervised approaches require large amounts of labelled data, which is expensive and time-consuming to obtain. This paper introduces a novel self-supervised video strategy for enhancing action prediction inspired by DINO (self-\textbf{di}stillation with \textbf{no} labels). The approach, named Temporal-DINO, employs two models; a `\student' processing past frames; and a `\teacher' processing both past and future frames, enabling a broader temporal context. During training, the \teacher~ guides the \student~ to learn future context by only observing past frames. The strategy is evaluated on ROAD dataset for the action prediction downstream task
using 3D-ResNet, Transformer, and LSTM architectures. The experimental results showcase significant improvements in prediction performance across these architectures, with our method achieving an average enhancement of 9.9\% Precision Points (PP), which highlights its effectiveness in enhancing the backbones' capabilities of capturing long-term dependencies. Furthermore, our approach demonstrates efficiency in terms of the pretraining dataset size and the number of epochs required. This method overcomes limitations present in other approaches, including the consideration of various backbone architectures, addressing multiple prediction horizons, reducing reliance on hand-crafted augmentations, and streamlining the pretraining process into a single stage. These findings highlight the potential of our approach in diverse video-based tasks such as activity recognition, motion planning, and scene understanding. Code can be found at \href{https://github.com/IzzeddinTeeti/ssl_pred}{https://github.com/IzzeddinTeeti/ssl\textunderscore{}pred}. 
\end{abstract}

\section{Introduction}
Computer vision techniques have advanced to the point at which they are able to outperform humans at certain object recognition tasks \cite{Elia2021}. However, for many computer vision applications, a higher-level understanding of the scene is required. For example, achieving human-level performance in autonomous vehicles remains a formidable challenge \cite{SHIWAKOTI2020870}. One of the key reasons for this gap is the inherent difficulty in understanding what may happen next. Thus there is a growing recognition of the importance of \textit{prediction}. Prediction plays a crucial role in enhancing the decision-making process of autonomous systems by anticipating the future behaviour of dynamic elements in the environment, \textit{e.g.} other vehicles, pedestrians, and cyclists - thus ensuring safer operation. Moreover, prediction also facilitates the development of high-level understanding within autonomous systems, enabling more nuanced and contextually appropriate planning, leading to smoother interactions with other agents on the road \cite{LI2022108499}.

However, prediction poses its own set of challenges, encompassing spatial, temporal, social, and stochastic dimensions \cite{teeti2022pred}. Modelling these dimensions requires complex models, such as \cite{Rasouli2019pie, Salzmann2020, Yuan2021, nayakanti2022wayformer}, which require significant amounts of data - often scarce and costly to gather and annotate. To address this, leveraging the abundance of unlabelled data through self-supervised methods offers an enticing opportunity to enhance performance with minimal impact upon resources. 

While existing self-supervised prediction methods, including \cite{tran2021knowledge, zatsarynna2022self, kochakarn2023explainable}, have shown promise, they have limitations that hinder their effectiveness. Firstly, their predictive capability is limited to a very short-term horizon (typically one frame ahead) which is impractical for autonomous driving scenarios requiring longer-term predictions. Secondly, these methods often involve a two-stage process \cite{tran2021knowledge}, which is computationally expensive and time-consuming. Finally, they are typically designed for a specific architecture, lacking the ability to generalize across different architectures. 

In this paper, we present a novel one-stage self-supervised representation learning strategy specifically designed for videos. Our proposed approach draws inspiration from the image-based self-supervised DINO \cite{caron2021emerging} model and extends its application to the temporal dimension. Leveraging a student-teacher framework, our method guides the \student~ model to focus on the most informative (temporal) features, enabling accurate predictions of future events. The \student~ model learns to attend to the relevant cues in the past and current moments, extracting valuable information that aids in forecasting forthcoming actions. Our approach addresses the aforementioned limitations by significantly extending the prediction range beyond one frame, enabling more practical and effective autonomous driving applications. Moreover, our proposed method eliminates the need for a two-stage process, reducing computational complexity and saving valuable time. Finally, our approach is not limited to a single architecture but is a wrapper strategy that can be used to improve prediction performance across various architectures, enhancing the generalisability and applicability of the method. The contributions of this work include:

\begin{itemize}
    \item A novel one-stage self-supervised representation learning strategy for videos, addressing limitations in prediction range, computational complexity, and architectural generalisability.
    \item Proof-of-concept evaluation of the approach in both autonomous driving, and a more general task - with various deep-learning architectures (including 3D-CNN, Transformer, and LSTM),   showcasing its effectiveness on real-world challenging data, and versatility to use on other models and domains.
    \item Identification of the optimal model architecture and loss function for capturing long-range dependencies in video data, shedding light on the most effective design choices for robust action prediction in autonomous driving scenarios.
\end{itemize}

In the following sections, we provide a comprehensive overview of related work in self-supervised learning for images and videos, and action prediction (Section \ref{sec:related}), followed by a detailed description of our proposed method (Section \ref{sec:method}). We then present the experimental setup, including datasets, evaluation metrics, and training details (Section \ref{sec:experiments}), and discuss the results and analysis of our experiments (Section \ref{sec:res}). Finally, we provide a thorough discussion of our findings, 
and highlight potential applications and future research directions (Section \ref{sec:conclusion}).

\section{Related Work}
\label{sec:related}

\subsection{Image-based Self-supervised Learning}
Two types of self-supervised strategies are commonly employed in computer vision: pretext methods and contrastive learning. The former involves utilizing specific internal properties or tasks of the input data to learn useful representations. For instance, some models learn the contextual information of the entire scene by analyzing small image patches \cite{doersch2015unsupervised}. Other approaches focus on tasks like re-colourization of grayscale images \cite{zhang2016colorful}, predicting transformations between different views of the same image \cite{gidaris2018unsupervised}, or determining the order of patches extracted from an image \cite{noroozi2016unsupervised, misra2020self}. With the introduction of Vision Transformer (ViT) \cite{dosovitskiy2020image}, patch-based self-supervised methods have gained prominence in the literature \cite{he2022masked, bao2021beit, assran2023self}. Particularly, masked auto-encoders (MAE) \cite{he2022masked} have emerged as a preferred mechanism for model pretraining due to their superior performance on downstream tasks. 

Contrastive learning methods, on the other hand, focus on maximizing the dissimilarity between features extracted from different samples to encourage discriminative representations \cite{dosovitskiy2014discriminative, bojanowski2017unsupervised, wu2018unsupervised}. Some discriminative self-supervised algorithms leverage instance-level discrimination to create distinct feature representations for different examples, leading to robust feature learning \cite{he2020momentum, chen2020simple, grill2020bootstrap, caron2021emerging}. Other approaches draw inspiration from clustering mechanisms \cite{caron2020unsupervised, asano2019self}. The DINO method \cite{caron2021emerging}, for example, employs self-distillation that trains a \student~ model on local crops and a \teacher~ model on the entire image, then find the loss between both representations. The \teacher~ will push the \student~ to learn global representations by seeing only the local ones. Whilst these techniques offer significant potential on images, their performance is limited on video-related tasks, such as action prediction. Thus, there is a need for advancements in video self-supervised learning methods.

\subsection{Video-based Self-supervised Learning}

Video-based self-supervised learning strategies, akin to their image counterparts, encompass both pretext and contrastive approaches \cite{schiappa2022self}. However, videos introduce an additional dimension, namely the temporal dimension. The inherent order of frames within a video sequence offers intrinsic properties that can be leveraged to develop effective self-supervised learning mechanisms. Despite this potential, the research in this specific domain remains relatively limited \cite{tong2022videomae, benaim2020speednet, misra2016shuffle, wang2015unsupervised}.

To address this gap, recent works have explored different strategies in video-based self-supervised learning. For instance, \cite{han2020self} employed complementary information from RGB and optical flow streams to co-train their model using contrastive loss. \cite{guo2022cross} utilized a 3D-CNN architecture with a video transformer and trained the model using contrastive loss on positive pairs of video sequences. CVRL \cite{qian2021spatiotemporal} employed a discriminative learning approach, utilizing two augmented views of the same video as positive examples and views of other videos as negatives. Another method, VideoMoCo \cite{pan2021videomoco}, extended the MoCo \cite{he2020momentum} framework to videos by randomly dropping frames from the video and learning the same representation for each random input. 

In line with these advancements, our proposed model introduces a novel approach where the \teacher~ model incorporates a longer temporal sequence than the \student~ model. This design choice aims to provide the \student~ with a wider temporal context, encompassing future frames that the \ student has not yet observed. By leveraging this extended temporal context within the student-teacher framework, our model aims to enhance the \student's ability to learn representations that capture long-term dependencies and improve its performance in video-based self-supervised learning.

\subsection{Supervised Action Prediction}

Solving the action prediction task requires modelling the different dimensions of the problem, including the spatial, temporal, stochastic, and social dimensions, since the driving environment is dynamic, uncertain, and multi-agent. To model the temporal dimension, \cite{Fang2021,Kotseruba2021} used 3D-CNN, \cite{Rasouli2019pie, girase2021loki} Recurrent Neural Networks, while \cite{Yuan2021} used Transformers. Regarding the stochastic dimension, GANs \cite{Kosaraju2019}, and CVAE models \cite{girase2021loki,Yuan2021} were utilised. To model the multi-agent aspect of the problem, different types of Graph Neural Networks of different connectivity, sparsity and homogeneity were used \cite{Gilles2021go, Salzmann2020, girase2021loki}, semantic segmentation cues \cite{Rasouli2021,Mangalam2021} and social pooling \cite{Pang2021}. However, all of those methods are supervised; they require a huge amount of labelled data which is time-consuming and expensive.

\subsection{Self-supervised Action Prediction}

In recent studies, limited approaches have been explored in the field of self-supervised action prediction. Zatsarynna et al. \cite{zatsarynna2022self} employed the contrastive loss of InfoNCE \cite{Gutmann2010} to encourage proximity between temporally adjacent video clips in the embedding space. To preserve the order of the clips, they complemented the InfoNCE loss with an order loss in the form of cross-entropy. Their proposed model focused on using 3D-CNN as the backbone architecture, and its evaluation was limited to this specific backbone without examining the performance on other architectures.

Another approach by Kochakarn et al. \cite{kochakarn2023explainable} utilized graph contrastive learning with the SimCLR loss function \cite{chen2020simple} to learn more informative embeddings for the prediction task. They incorporated an attention mechanism to achieve explainable action prediction, albeit limited to predicting only the next one frame. In contrast, our proposed method extends the prediction horizon to include the next 3, 6, and 12 frames, offering a more comprehensive temporal context. Furthermore, their approach specifically focused on graphs, while our method is designed to work with 3D-CNN, Transformers, and LSTMs, providing broader applicability.

It is worth noting that both of these aforementioned approaches rely heavily on the use of contrastive loss, which requires careful crafting of augmentations. This dependency on specific augmentations can limit the generalisability and robustness of the learned representations.

In contrast, Tran et al. \cite{tran2021knowledge} adopted a knowledge distillation approach, where a \teacher~ model trained on recognition tasks transfers its knowledge to a prediction model. However, this method follows a two-stage process involving training a \teacher~ model for recognition and then distilling the knowledge into a \student~ model for prediction. This two-stage approach introduces additional time and computational costs, which may impact the scalability and practicality of the method, particularly in real-time or resource-constrained scenarios. Furthermore, their evaluation was focused on the I3D architecture \cite{carreira2017quo} without exploring the performance on other architectures.

These related works provide valuable insights into self-supervised action prediction approaches. However, they have certain limitations in terms of the backbone architectures considered, the prediction horizons addressed, the reliance on specific augmentations, and the two-stage training process. In contrast, our one-stage proposed method aims to overcome these limitations by leveraging a different loss formulation and considering multiple backbone architectures while extending the prediction horizon, thereby contributing to the advancement of self-supervised action prediction techniques.

\section{Methodology}
\label{sec:method}

\subsection{Representation Learning for Action Prediction}

In this study, our objective is to learn a mapping function $f$ in an unsupervised manner, which takes an unlabelled and untrimmed video clip consisting of $t \in \mathbb{R}^{+}$ frames as input. The goal is to map the 4D input clip $x_{1:t} \in \mathbb{R}^{T{\times}C{\times}H{\times}W}$, to a feature vector $g(x_{1:t}) \in \mathbb{R}^{d}$, such that the learned features effectively transfer to the downstream task of action prediction. To achieve this, we draw inspiration from the DINO approach and adopt a student-teacher setup, as depicted in Figure \ref{fig:overview}.

The \student~ network $S_{\theta}$ processes only the past frames $x_{1:t}$ during both training and inference, without access to future frames. On the other hand, the \teacher~ network $T_{\phi}$ processes both the past and future frames $x_{1:t+t_{Pred}}$, where $t_{Pred}$ denotes the length of the future sequence. To ensure consistency in architecture, we downsample the sequence processed by the \teacher~ network to match the number of frames processed by the \student. The downsampling is performed by determining a sampling frequency, which is calculated as $\nicefrac{(t+t_{Pred})}{t}$. For instance, if $t=12$ frames and $t_{Pred}=12$ frames, the \student~ processes the past (12) frames, while the \teacher~ processes (24 frames) with a step of 2, resulting in 12 frames. Despite processing sequences of the same sequence length, the \teacher~ network has access to a wider temporal context.

\subsection{Future-past Distillation Loss}

During training, we introduce a knowledge distillation loss between the final embeddings of the \student~ and \teacher~ networks. This loss guides the \student~ to distil knowledge about the future from the \teacher, despite not having direct access to future frames. The aim is to teach the \student~ to focus on the most relevant features from the past frames that contribute to predicting the future. In contrast to DINO, our Future-Past Distillation (FPD) loss is defined in the Cosine Similarity form, instead of using cross-entropy. This formulation is motivated by the findings of our ablation analysis, which indicate that the Cosine-based FPD loss yields improved performance on downstream tasks. The learning objective for the pretraining stage of future-past distillation is expressed in Equation \ref{eqn:fpd_min}. The \student~ network parameters $\theta$ are updated using backpropagation optimized by stochastic gradient descent (SGD), while the \teacher~ network parameters $\phi$ are updated using an Exponential Moving Average (EMA) based on the \student~ network, with a scheduled momentum variable ($m$) as shown in Equation \ref{eqn:ema}.

\vspace{-0.2cm}
\begin{equation}
    \label{eqn:fpd_min}
    \theta^{*}, \phi^{*} = \argmin_{\theta, \phi} \mathcal{L}_{FPD}\Big(S_{\theta}(x_{1:t}), T_{\phi}(x_{1:t+t_{Pred}})\Big)
\end{equation}


\begin{equation}
    \phi_{i+1} = m_{i} \times \phi_{i} + (1-m_{i}) \times \theta_{i}
    \label{eqn:ema}
\end{equation}

\subsection{Downstream Task Definition}

The objective of pretraining is to enhance the performance of the model in the downstream task of predicting driver's actions. Given a past (observed) clip of length $t \in \mathbb{R}^{+}$, the task is to predict the Ego-vehicle (driver) action in each frame in the next $t_{Pred} \in \mathbb{R}^{+}$ frames. 

Building upon the optimal pretrained \student~ model $S_{\theta^{*}}$, the prediction model $f$ will use the \student~ model as a backbone and add a classification head on top of it. Subsequently, it performs further optimization (fine-tuning) on either both the backbone and the head parameters or solely the latter (we conducted experiments on both scenarios) for the prediction task. The objective is to map the learned features $S_{\theta^{*}}(X_{1:t})$ to the future action labels $y_{t+1:t+t_{Pred}}$, formally, $f_{\psi}:S_{\theta^{*}}(X_{1:t}) \rightarrow y_{t+1:t+t_{Pred}}$. Given the nature of a classification task, cross-entropy (CE) loss is utilized to guide the optimization process and refine the learning objective for the downstream task, as illustrated in Equation \ref{eqn:pred}.

\begin{equation}
    \label{eqn:pred}
    \theta^{**}, \psi^{*} = \argmin_{\theta^{*}, \psi} \mathcal{L}_{CE}\bigg(f_{\psi}\Big(S_{\theta^{*}}(x_{1:t})\Big), y_{t+1:t+t_{Pred}}\bigg)
\end{equation}

\section{Experiments}
\label{sec:experiments}

\subsection{Datasets}
\label{sec:datasets}

We adhered to the convention in self-supervised learning, utilizing two distinct datasets for distinct purposes: a larger dataset for pretext task pretraining and a smaller dataset for downstream task fine-tuning. Specifically, we employed the Kinetics-400 dataset \cite{kay2017kinetics} and the ROad event Awareness Dataset (ROAD) \cite{SinghRoad}.

\paragraph{Kinetics-400} 
It is designed for action recognition and comprises over 240,000 videos. On average, each video spans 10 seconds and is assigned a single label from a pool of 400 possible action classes. The dataset's substantial video collection has facilitated its adoption in numerous video self-supervised methods \cite{tong2022videomae, pan2021videomoco}. For our purposes, we disregard the label information as it will not be used during pretraining.

\paragraph{ROAD}
The ROad event Awareness Dataset (ROAD) \cite{SinghRoad} is built on a fraction of Oxford RobotCar Dataset \cite{maddern20171}, and it is extended with multi-label annotations for action recognition, localisation, and prediction tasks within the context of autonomous driving. It comprises 22 videos from an ego-centric view as shown in Figure \ref{fig:road}, each with an 8-minute duration and a frame rate of 12 frames per second (fps). Importantly, it contains labels indicating the actions performed by the ego vehicle (driver). Notably, the dataset encompasses seven distinct ego-vehicle actions: \textit{Move, Stop, Turn Left, Turn Right, Overtake, Move Left, and Move Right}. These labels serve as training data for fine-tuning our models to predict the driver's actions in future frames during the prediction task.


\subsection{Models}
\label{sec:models}

We conducted a series of experiments utilizing four diverse deep-learning architectures. 

\vspace{-0.2cm}
\paragraph{R3D \cite{D3ResNet2017}} This architecture employs a 3D convolutional neural network (3D-CNN) backbone for processing video data. It is based on the ResNet-18 \cite{he2016deep} architecture, which has proven to be successful in image recognition tasks. 

\vspace{-0.2cm}
\paragraph{Swin \cite{liu2021swin}} This architecture utilizes a Transformer-based 3D backbone for video processing. It is based on the Transformer architecture, originally introduced in the context of image recognition as Vision Transformer (ViT) \cite{dosovitskiy2020image}. 

\vspace{-0.2cm}
\paragraph{ResNet-LSTM} This architecture combines a convolutional neural network (CNN)-based 2D backbone using the ResNet-50 architecture for image processing, along with a Long Short-Term Memory (LSTM) layer for capturing temporal dependencies. 

\vspace{-0.2cm}
\paragraph{ViT-LSTM} Same as the previous but replacing the ResNet with a Transfomer-based 2D backbone.  

The first two models use spatio-temporal backbones, which extract spatial and temporal features simultaneously, while the last two use a spatial backbone and connect it using a temporal one. These models exhibit differences in their depth and working mechanisms, offering a diverse range of approaches to our experiments. By leveraging these varied architectures, we can comprehensively investigate our proposed representation learning strategy, examine their performance, and compare their effectiveness according to the experiments outlined in the subsequent section.

\subsection{Experimental Protocol}
\label{sec:exp_protocol}

In accordance with the conventions of self-supervised learning experiments, we evaluated the performance of our models on the downstream task of action prediction under three distinct protocols: \\1) \textit{Full-Supervised:} This protocol represents the results obtained from the model without employing the pretraining strategy. The model was trained solely on the labelled data of the prediction task. \\2) \textit{Linear Probing:} In this protocol, the proposed strategy was applied, and the model was fine-tuned by solely updating the parameters of the prediction head. The backbone of the model was kept frozen during this fine-tuning process. \\3) \textit{Fine-tuning:} This protocol involved applying the proposed strategy and performing full fine-tuning of the entire network, including both the backbone and the prediction head. All parameters of the model were updated during the fine-tuning process.

By examining the model's performance across these three protocols, we can gain insights into the effectiveness and impact of the pretraining strategy on action prediction.

\subsection{Implementation Details}
We used Pytorch framework \cite{paszke2019pytorch} to implement the models, and the Precision (P) metric was used to evaluate their performance on the action prediction task. All experiments were conducted on NVIDIA A30 graphics cards. Below are the details of the pretraining and fine-tuning.

\paragraph{Pretraining}
For pretraining, we utilized either the full dataset of Kinetics-400 or the training split of ROAD (depending on the experiment). The pretraining was optimised for 1000 (50) epochs using SGD optimiser with a learning rate of 0.005 (0.001) and a batch size of 64 (32) for Kinetics-400 and ROAD, respectively. Additionally, a cosine scheduler was employed for updating the momentum variable of the exponential moving average (EMA). Concerning the models, the LSTM architecture had a hidden dimension of 512, and the small structure of ViT (ViT-s) was used.

\paragraph{Fine-tuning}
Fine-tuning was performed on the ROAD dataset, with a data split of 60\% for training, 20\% for validation, and 20\% for testing. Cross-entropy loss was employed as the objective function for fine-tuning, and it was optimised for 10 epochs using the SGD optimizer with a learning rate of 0.001, and a batch size of 32.

\begin{figure}
    \centering
    \subfloat[ROAD dataset]{\includegraphics[width=0.45\textwidth]{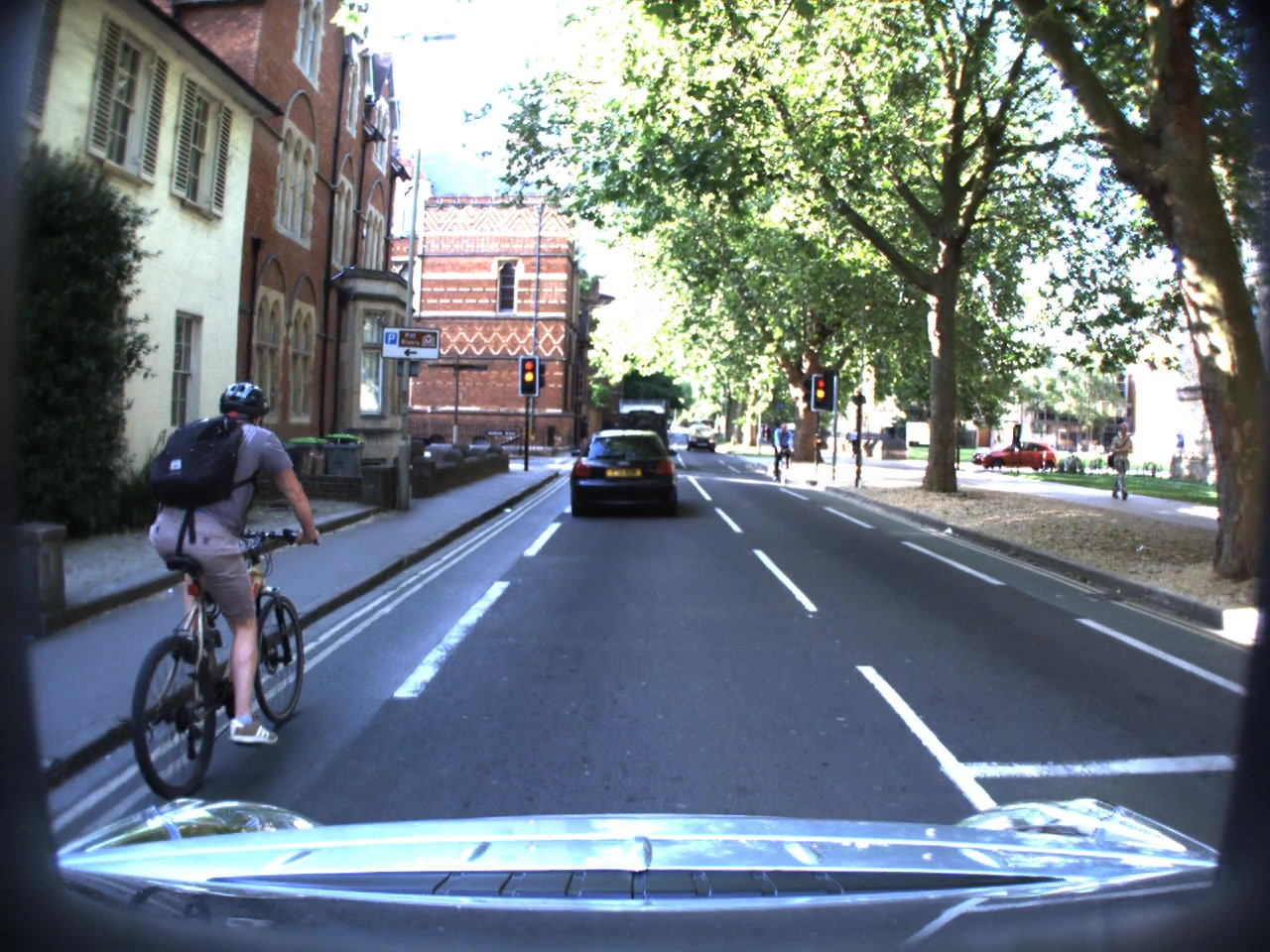}} \hspace{5mm}
    \subfloat[UCF101 dataset]{\includegraphics[width=0.45\textwidth]{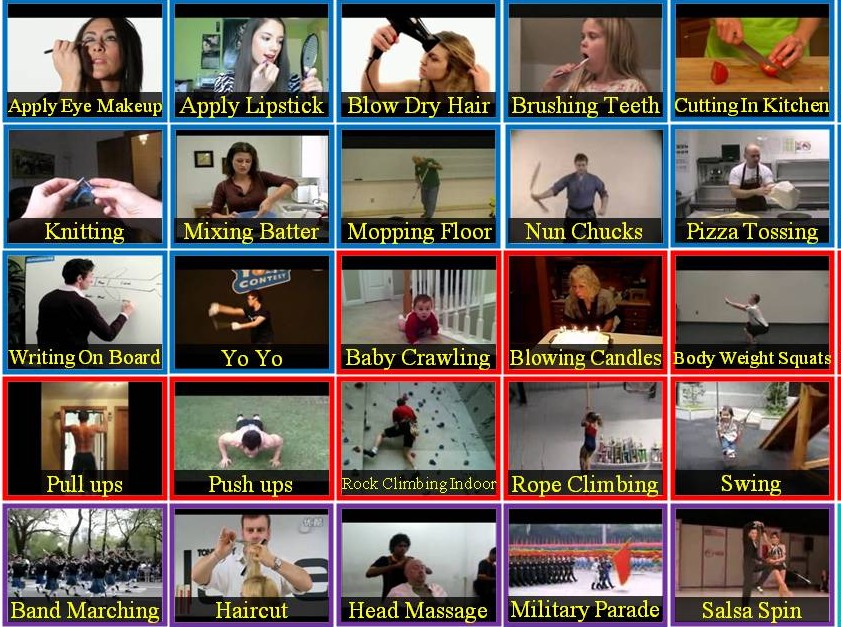}}
    
    \caption{Sample images from ROAD and UCF101 datasets.} 
    \label{fig:road}
\end{figure}

\subsection{Ablations}

In this section, we present different ablations to study the effects of different architectural components on the model performance. Specifically, we investigate the effects of variations in input sequence temporal length, self-supervised objective, backbone selection, and 
the strategy's performance on another downstream task (action recognition). Detailed explanations of each ablation study are provided below.

\paragraph{Backbone and Temporal Length}
The choice of model backbone plays a critical role in the model's ability to learn effective features. Similarly, the length of the input sequence contributes to the model's capability to capture temporal dynamics and visual changes within the scene. Longer sequence lengths generally provide more visual data, but they may also introduce challenges in modelling long-term dependencies and potentially lead to performance degradation, as shown in Table \ref{tab:interval_backbone}. 

In this specific ablation study, We conducted experiments with the four backbones mentioned in Section \ref{sec:models}, and we varied the temporal depth by using sequence lengths of 3, 6, and 12 frames for each backbone. We trained these models using the three experimental protocols outlined in Section \ref{sec:exp_protocol}. The prediction performance of the four models with different input lengths under the three protocols is summarized in Table \ref{tab:interval_backbone}.

\begin{table*}[h]
\centering
\caption{The precision of different backbones with varying input lengths under the three protocols mentioned in Section \ref{sec:exp_protocol}.}
\label{tab:interval_backbone}
\begin{tabular}{@{}ccccccc@{}}
\toprule
Backbone &
  \begin{tabular}[c]{@{}c@{}}Pretrained\\  on\end{tabular} &
  \begin{tabular}[c]{@{}c@{}}Interval\\ (frames)\end{tabular} &
  \begin{tabular}[c]{@{}c@{}}Linear\\ Probe\end{tabular} &
  Fine-tunning &
  Supervised &
  Improvement \\ \midrule
\multirow{3}{*}{R3D}         & \multirow{3}{*}{ROAD}         & 3  & 36.6  & 77.7 & 74.1 & 3.6  \\
                             &                               & 6  & 29.7  & 69.2 & 64.9 & 4.3  \\
                             &                               & 12 & 36.2  & 53.3 & 45.7 & 7.6  \\ \midrule
\multirow{3}{*}{Swin}        & \multirow{3}{*}{ROAD}         & 3  & 84.7  & 87.2 & 86.4 & 0.8  \\
                             &                               & 6  & 75.8  & 82.6 & 81.7 & 0.9  \\
                             &                               & 12 & 60.1  & 60.9 & 57.1 & 3.8  \\ \midrule
\multirow{3}{*}{ResNet+LSTM} & \multirow{3}{*}{Kinetics-400} & 3  & 77.6  & 84.6 & 62.9 & 21.7 \\
                             &                               & 6  & 70.3  & 81.8 & 58.3 & 23.5 \\
                             &                               & 12 & 58.3  & 76.7 & 54.3 & 22.4 \\ \midrule
\multirow{3}{*}{ViT+LSTM}    & \multirow{3}{*}{Kinetics-400} & 3  & 73.5  & 77.7 & 69.3 & 8.4  \\
                             &                               & 6  & 70.2  & 76.7 & 65.5 & 11.2 \\
                             &                               & 12 & 64.21 & 66.2 & 55.5 & 10.7 \\ \bottomrule
\end{tabular}
\end{table*}

\paragraph{Loss Function}
Existing literature demonstrates that the choice of learning objective in self-supervised algorithms significantly influences the performance of models on downstream tasks \cite{schiappa2022self}. In our experiments, we employed three widely used loss functions: Cross-entropy, Cosine Similarity, and Mean Squared Error (MSE) loss. The evaluation of these loss functions was performed on R3D and Swin backbones, and the corresponding results are summarized in Table \ref{tab:loss}.

\begin{table*}[h]
\centering
\caption{Evaluation of three common loss functions on R3D and Swin backbones.}
\label{tab:loss}
\begin{tabular}{@{}cccccc@{}}
\toprule
Backbone              & Loss          & Linear Probe & Fine-tunning & Supervised & Improvment \\ \midrule
\multirow{3}{*}{R3D}  & MSE           & 32.0            & 37.3            & 45.7          & -8.4          \\
                      & Cosine        & 36.2            & \textbf{53.3}            & 45.7          & 7.6          \\
                      & Cross-entropy & 30.9            & 50.4            & 45.7          & 4.7          \\ \midrule
\multirow{3}{*}{Swin} & MSE           & 57.2            & 57.4            & 57.1          & 0.3          \\
                      & Cosine        & 60.1            & \textbf{60.9}            & 57.1          & 3.8          \\
                      & Cross-entropy & 49.6            & 58.4            & 57.1          & 1.3          \\ \bottomrule
\end{tabular}
\end{table*}

\paragraph{Action Recognition}
In addition, we performed supplementary experiments to assess the impact of our proposed strategy on a different video-based downstream task, namely Action Recognition. For this purpose, the models underwent pretraining on the Kinetics-400 dataset and subsequent fine-tuning on the UCF101 dataset \cite{soomro2012ucf101}, shown in Figure \ref{fig:road}. Table \ref{tab:sota_ucf} presents the state-of-the-art (SOTA) performance for action recognition.

\section{Results}
\label{sec:res}

Observing Table \ref{tab:interval_backbone}, it is evident that our proposed strategy yielded notable enhancements in the prediction performance of all backbone architectures, albeit to varying extents. Notably, the impact of the strategy is particularly pronounced on the 2D-based backbones, which initially exhibited comparatively lower results in the fully-supervised setup. This observation aligns with expectations since these backbones lack inherent spatio-temporal modelling capabilities. However, when trained with T-DINO, the ResNet-LSTM and ViT-LSTM backbones witnessed substantial improvements, with average increases of 22.5 and 10.1 in terms of precision points (PP), respectively. Additionally, the video backbones, R3D and Swin, experienced PP gains of 5.2 and 1.8, respectively. Table \ref{tab:sota_road} shows the comparison of our strategy to other SOTA methods, showing that it surpasses the performance of both supervised and self-supervised approaches on action prediction on ROAD. Within the same table, the column `Supervised' indicates a fully supervised method, the opposite denotes a self-supervised approach, and `Frozen' means the finetuning process exclusively updates the classification head while leaving the pre-trained backbone untouched, the opposite indicates that the entire model was updated during the fine-tunning.

\begin{table*}[]
\centering
\caption{Comparison of the proposed methods with SOTA on ROAD for the action prediction task when the input length is 12 frames}
\label{tab:sota_road}
\begin{tabular}{lccccc}
\hline
Method                                     & Pretrained on & Backbone    & Supervised            & Frozen                & Precision           \\ \hline
R3D \cite{tran2018closer} & Road          & R3D         & \cmark & \xmark & 45.7          \\
Swin \cite{liu2022video}  & Road          & Swin        & \cmark & \xmark & 57.1          \\
Resnet+LSTM \cite{yue2015beyond} & Road     & ResNet50 & \cmark & \xmark & 54.3       \\
ViT+LSTM                                   & Road          & ViT         & \cmark & \xmark & 55.5          \\
VideoMAE \cite{tong2022videomae} & Kinetics & ViT      & \xmark & \xmark & {\ul 75.1} \\ \hline
T-DINO (ours)                              & Road          & R3D         & \xmark & \cmark & 53.3          \\
T-DINO (ours)                              & Kinetics      & ResNet+LSTM & \xmark & \cmark & 58.3          \\
T-DINO (ours)                              & Kinetics      & ResNet+LSTM & \xmark & \xmark & \textbf{76.7} \\
T-DINO (ours)                              & Kinetics      & ViT+LSTM    & \xmark & \cmark & 64.2          \\
T-DINO (ours)                              & Kinetics      & ViT+LSTM    & \xmark & \xmark & 66.2          \\ \hline
\end{tabular}
\end{table*}

To have a better understanding of the generalisation capability of the proposed model, we compare the performance of T-DINO with SOTA methods on another task, human action recognition (on UCF101), summarised in Table \ref{tab:sota_ucf}. The results highlight the effectiveness of the enhanced temporal modelling offered by T-DINO when applied to the R3D backbone, 

\begin{table*}[h]
  \caption{Comparison of the proposed methods with SOTA on UCF101 for action recognition task.}
  \label{tab:sota_ucf}
  \centering
  \begin{tabular}{lcccccc}
    \toprule
    Method & Year & Pretrained on & Arch. & Supervised & Frozen & Acc. \\
    \hline
    ClipOrder \cite{xu2019self} & 2019 & UCF101 & R3D & \xmark & \xmark & 72.40 \\
    3D ST-puzzle \cite{kim2019self} & 2019 & Kinetics & C3D & \xmark & \xmark & 65.80 \\
    Wang et al. \cite{wang2019self} & 2019 & UCF101 & C3D & \xmark & \xmark & 61.20 \\
    PRP \cite{yao2020video} & 2020 & Kinetics & R3D & \xmark & \xmark & 72.10 \\
    SpeedNet \cite{benaim2020speednet} & 2020 & Kinetics &  S3D-G & \xmark & \xmark & 81.10 \\
    CSJ \cite{huo2021self} & 2021 & K+UCF101 & R(2+3)D & \xmark & \xmark & 79.50 \\
    VideoMoCo \cite{pan2021videomoco} & 2021 & Kinetics & R(2+1)D & \xmark & \xmark & 78.7 \\
    CACL \cite{guo2022cross} & 2022 & UCF101 & R(2+1)D & \xmark & \xmark & 82.5 \\
    VideoMAE \cite{tong2022videomae} & 2022 & Kinetics & ViT & \xmark & \xmark & \textbf{96.1} \\
    \hline
    T-DINO (ours) & 2023 & Kinetics & ResNet+LSTM & \xmark & \xmark & 80.26 \\
    T-DINO (ours) & 2023 & Kinetics & R3D & \xmark & \xmark & 83.9 \\
    T-DINO (ours) & 2023 & Kinetics & ViT+LSTM & \xmark & \xmark & {\ul85.86} \\
    \bottomrule
  \end{tabular}
\end{table*}

Furthermore, examining the results for varying input sequence intervals across each backbone, it becomes apparent that greater improvements are observed at longer input sequences. This suggests that pretraining with T-DINO equips the backbones with enhanced abilities to capture and model long-term dependencies in the data. Notably, the Swin-transformer-based models demonstrate higher accuracy, attributed to the superior representation capabilities offered by Transformers. We observed that transformer-based spatial feature extractor (ViT) combined with LSTM-based temporal sequence modelling results in a huge improvement in the downstream task. Analyzing the results obtained from the longest input configuration consisting of 12 frames, it is evident that models pretrained on the larger dataset, Kinetics-400, exhibit superior performance compared to those pretrained on the ROAD dataset. Moreover, the models that incorporate separate modelling of spatial and temporal relationships, such as ViT+LSTM and ResNet+LSTM, outperform the models that jointly model these relationships, namely R3D and Swin, by a margin of 16.6 and 5.7 percentage points, respectively.

In regard to the selection of the most optimal loss function, the findings presented in Table \ref{tab:loss} indicate that T-DINO pretrained using the Cosine Similarity loss surpasses the performance of models trained with MSE or Cross-entropy loss.

Of significant importance, self-supervised models typically require extensive datasets and a high number of pretraining epochs to achieve satisfactory generalization on downstream tasks. However, our proposed strategy, pretrained on the ROAD dataset, exhibits a relatively comparable level of performance to models pretrained on the larger Kinetics-400 dataset. Notably, the ROAD dataset possessed a substantially smaller size and was pretrained with a lower number of epochs. These findings demonstrate T-DINO's resource and time efficiency in achieving desirable results.

\section{Conclusion}
\label{sec:conclusion}

This study represents the first attempt to leverage future information in a `past training' model, and the promising results indicate that this teacher-student approach could provide a significant performance improvement in various prediction tasks across a number of ubiquitous model architectures in a variety of different domains - without the requirement for any additional training data. Our proposed strategy, called Temporal-DINO, leverages a teacher-student self-distillation architecture to guide the student model to learn future temporal context by observing the past only. Unlike other approaches that involve a two-stage process or rely on hand-crafted augmentations with limited prediction horizons, our one-stage strategy overcomes these limitations. Additionally, ablations highlight the strategy's generalisability, efficiency, and feasibility in hardware-constrained applications.

In terms of future directions, several avenues can be pursued to further enhance and expand our proposed approach. Firstly, the inclusion of Graph Neural Networks (GNNs) as additional architectural variations could be explored to examine the strategy's ability to enhance the social dimension modelling. Secondly, expanding the evaluation of our approach to encompass a broader range of datasets from diverse domains.

{\small
\bibliographystyle{ieee_fullname}
\bibliography{egbib}
}

\end{document}